# Artificial Ant Colonies in Digital Image *Habitats* – A Mass Behaviour Effect Study on Pattern Recognition

Vitorino Ramos[1], Filipe Almeida[2]

[1]CVRM - IST Geo-Systems Center, Instituto Superior Técnico,
Avenida Rovisco Pais, 1049-001, Lisboa, Portugal
`vitorino.ramos@alfa.ist.utl.pt`

[2]VARIOGRAMA.com, Largo do Corpo Santo,
nº6 – 1º, 1200-129, Lisboa, Portugal
`fra@mega.rnl.ist.utl.pt`

## Abstract

Some recent studies have pointed that , the self-organization of neurons into brain-like structures, and the self-organization of ants into a swarm are similar in many respects. If possible to implement, these features could lead to important developments in pattern recognition systems, where perceptive capabilities can emerge and evolve from the interaction of many simple local rules. The principle of the method is inspired by the work of *Chialvo* and *Millonas* who developed the first numerical simulation in which swarm cognitive map formation could be explained. From this point, an extended model is presented in order to deal with digital image *habitats*, in which artificial ants could be able to react to the environment and perceive it. Evolution of pheromone fields point that artificial ant colonies could react and adapt appropriately to any type of digital habitat.

## 1. From Natural to Digital Habitats

In "*Godel, Escher, Bach*", *Douglas Hofstadter* explores the difference between an ant colony as a whole and the individual that compose it. According to *Hofstadter*, the behaviour of the whole colony is far more sophisticated and of very different character than the behaviour of the individual ants. A colony's collective behaviour exceeds the sum of its individual member's actions (so-called *emergence*) and is most easily observed when studying their foraging activity. Most species of ants forage collectively using chemical recruitment strategies, termed *pheromone* trails, to lead their fellow nest-mates to food sources. This analogy with the way that real and natural ant colonies work and migrate, has suggested the definition in 1991/92 of a new computational paradigm, which is called the *Ant System* by *Marco Dorigo* [8]. In these studies there is *no precommitment* to any particular representational scheme: the desired behaviour is specified, but there is minimal specification of the mechanism required to generate that behaviour, i.e. *global behaviour evolves from the many relations of multiple simple behaviours*. Since then several studies were conducted to apply this recent paradigm – or analogous ones - in real case problems, with successful results. The new heuristic has the following desirable characteristics: (1) It is *versatile*, in that it can be applied to similar versions of the same problem; (2) It is *Robust*. It can be applied with only minimal changes to other problems (e.g. combinatorial optimisation problems such as the quadratic assignment problem - QAP, travelling salesman problem - TSP, or the job-shop scheduling problem - JSP);…and (3) It is a *population based approach*. This last property is interesting since it allows the exploitation of positive feedback as a search mechanism (the collective behaviour that emerges is a form of *autocatalytic* "snow ball" effect - that reinforces itself - behaviour, where the more the ants follow a trail, the more attractive that trail becomes for being followed). It also makes the system amenable to parallel implementations (though, only the intrinsically parallel and distributed nature of these systems are considered in this work). However, none of these new paradigms has been applied to Pattern Recognition problems (and namely to the sub-problem of image segmentation). The application of these heuristics on image segmentation looks very promising, since segmentation can be looked as a clustering and combinatorial problem [19, 20], and grey level image itself as a topographic map (where the image is the ant colony *habit*). Moreover, and as reported by some studies, the self-organisation of neurones into brain-like structures, and the self-organisation of ants into a swarm are similar in many respects. The present paper is based on *Chialvo* and *Millonas* work [4], which is probably, one of the simplest (local, memoryless, homogeneous and isotropic) models which leads to trail forming, where the formation trails and networks of ant traffic are not imposed by any special boundary conditions, lattice topology, or additional behavioural rules. The next section is dedicated to explain how this model operate in a finite square lattice, while the remaining sections will be devoted to extend their work to digital image lattices, where the respective grey level intensities will be considered in each ant *perception* of his neighbourhood. The main goal is to achieve a global perception of one image as the emergent sum of local perceptions of the whole colony.

## 2. The *Chialvo* and *Millonas* Swarm Cognitive Map Model

As described by *Chialvo* and *Millonas* in [4], the state of an individual ant can be expressed by its position $r$, and orientation $\theta$. Since the response at a given time is assumed to be independent of the previous history of the individual, it is sufficient to specify a transition probability from one place and orientation $(r,\theta)$ to the next $(r^*,\theta^*)$ an instant later. In previous works [16,17] transition rules were derived and generalized from noisy response functions, which in turn were found to reproduce a number of experimental results with real ants. The response function can effectively be translated into a two-parameter transition rule between the cells by use of a pheromone weigthing function:

$$W(\sigma) = \left(1 + \frac{\sigma}{1 + \delta\sigma}\right)^\beta$$

This equation measures the relative probabilities of moving to a cite $r$ (in our context, to a pixel) with pheromone density $\sigma(r)$ as discussed in [16,17]. The parameter $\beta$ is associated with the osmotropotaxic sensitivity, recognised by *Wilson* [25] as one of two fundamental different types of ants sense-data processing. *Osmotropotaxis*, is related to a kind of instantaneous pheromonal gradient following, while the other, *klinotaxis*, to a sequential method (though only the former will be considered in the present work as in [4]). Also it can be seen as a physiological inverse-noise parameter or gain. In practical terms, this parameter controls the degree of randomness with wich each ant follows the gradient of pheromone. As putted by both authors, for low values of $\beta$ the pheromone concentration does not greatly affect its choice, while high values cause it to follow pheromone gradient with more certainty. On the other hand, $1/\delta$ is the sensory capacity, which describes the fact that each ant's ability to sense pheromone decreases somewhat at high concentrations. In addition to the former equation, there is a weigthing factor $w(\Delta\theta)$, where $\Delta\theta$ is the change in direction at each time step, i.e. measures the magnitude of the difference in orientation. This weighting factor ensures that very sharp turns are much less likely than turns through smaller angles; thus each ant in the colony have a probabilistic bias in the forward direction. For instance, if we consider that one agent is coming from the north direction (fig. 1a), south-west (fig. 1b), or north-east (fig.1c), the following values $w(\Delta\theta)$ will be considered:

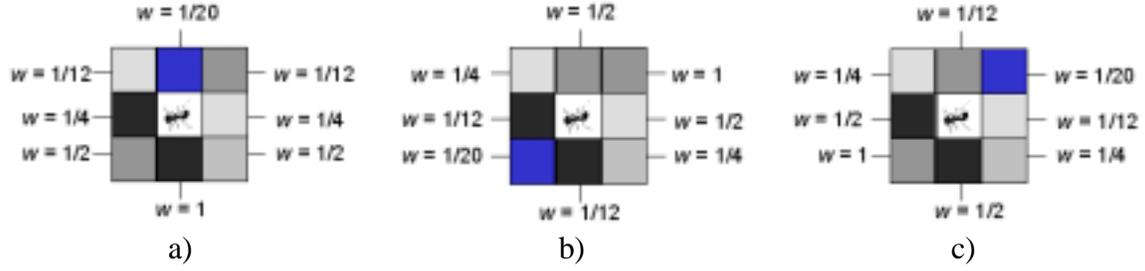

Fig.1 - Values for probabilist directional bias $w(\Delta\theta)$, for three examples. a) One ant is coming from north, b) from south-west, and one coming from north-east (values from [4]).

To be able to translate the former behaviour into a as real as possible coherent behaviour some discretization of space and time is needed. One possible discretization is to allow each ant to move from one cite to another on a square lattice. This last condition (though others are possible) not only follows the proposed model in [4], as it is the most appropriate for digital image implementations. As a consequence of discretizing the time at the ant colony, each individual can take one step at each time step, and similarly for space discretizations, each ant at each time step finds itself in one of these pixels, being its sensory input influenced by the concentration of pheromone in all the eight neighbouring cells. As an additional condition, each individual leaves a constant amount $\eta$ of pheromone at the pixel in which it is located at every time step $t$. This pheromone decays at each time step at a rate $k$. As in [4], toroidial boundary conditions are imposed on the lattice to remove, as far as possible any boundary effects (e.g. one ant going out of the image at the south-west corner, will probably come in at the north-east corner). Then, the normalised transition probabilities on the lattice to go from cell $k$ to cell $i$ are given by [4]:

$$P_{ik} = \frac{W(\sigma_i)w(\Delta_i)}{\sum_{j/k} W(\sigma_j)w(\Delta_j)}$$

where the notation $j/k$ indicates the sum over all the pixels $j$ which are in the local neighbourhood of $k$. $\Delta_i$ measures the magnitude of the difference in orientation for the previous direction at time $t$-1. That is, since we use a neighbourhood composed of the cell and its eight neighbours, $\Delta_i$ can take the discrete values 0 through 4, and it is sufficient to assign a value $w_i$ for each of these changes of direction. *Chialvo* and *Millonas* used the weights of (same direction) $w_0 = 1$, $w_1 = 1/2$, $w_2 = 1/4$, $w_3 = 1/12$ and $w_4 = 1/20$ (U-turn, see fig.1). Once these parameters are set, a large number of ants can be place on the image at random positions for time $t=0$. Then, for future computations, the random movement of each ant is determined by the probabilities $P_{ik}$. *Chialvo* and *Millonas* used a number of individuals equal to about 30% ($n=0.3$) of the squared lattice global area, value that will be used in the present work as a default value. For the swarm model to be complete, some conditions must be described for how pheromone evolves on the habitat; i.e. from which emergent learning will be possible (in fact, and in some sense a kind of reinforcement learning). Both authors [4], assumed that each organism emits pheromone at a given rate $\eta$, and that this quantity remains fixed at the emission point (i.e. there is no spatial diffusion). As the pheromone evaporates at rate $k$, the pheromonal field will contain information about past movements of the organisms, but not arbitrarily in the past, since the field *forgets* its distant history due to evaporation in a time $\tau \cong 1/k$. As mentioned by the authors, the distribution of the pheromone represents the memory of the recent history of the swarm, and in a sense it contain information which the individual ants are unable to hold or transmit. Note that there is no direct communication between the organisms but a type of indirect communication through the pheromonal field. In fact, ants are not allowed to have any memory and the individual's spatial knowledge is restricted to local information about the whole colony pheromone density.

Particularly interesting for the present work, and as defended in [4], the self-organisation of ants into a swarm and the self-organisation of neurones into a brain-like structure are similar in many respects. Swarms of social insects construct trails and networks of regular traffic via a process of pheromone laying and following. These patterns constitute what is known in brain science as a cognitive map. The main differences lies in the fact that insects write their spatial memories in the environment, while the mammalian cognitive map lies inside the brain, a fact that also constitutes an important advantage in the present extended model. As mentioned by the two authors, this analogy can be more than a poetic image, and can be further justified by a direct comparison with the neural processes associated with the construction of cognitive maps in the hippocampus. In [25], *Wilson* forecasted the eventual appearance of what he called "a stochastic theory of mass behaviour" and asserted that "the reconstruction of mass behaviours from the behaviours of single colony members is the central problem of insect sociobiology". He forecasted that our understanding of individual insect behaviour together with the sophistication with which we will able to analyse their collective interaction would advance to the point were we would one day posses a detailed, even quantitative, understanding of how individual "probability matrices" would lead to mass action at the level of the colony. As stated in [4], by replacing *colony members* with *neurones*, *mass behaviours* or *colony* by *brain behaviour*, and *insect sociobiology* with *brain science* the above paragraph could describe the paradigm shifts in the last twenty-five years of progress in the brain sciences [25]. In addition, coherent results were found for $\eta=0.07$ (pheromone deposition rate), $k=0.015$ (pheromone evaporation rate), $\beta=3.5$ (osmotropotaxic sensitivity) and $\delta=0.2$ (inverse of sensory capacity), where the emergence of well defined networks of trails were possible. For a detailed mathematical discussion of this model, readers are also reported to [4]. Except when indicated, these values will remain in the following image analysis oriented framework.

## 3. From Cognitive Maps to Perceptual Grouping *Gestalt* Fields

Image segmentation is a low-level image processing task that aims at partitioning an image into homogeneous regions [9]. How region homogeneity is defined depends largely on the application. A great number of segmentation methods are available in the literature to segment images according to various criteria such as for example grey level, colour, or texture. This task is hard and as we know very important, since the output of an image segmentation algorithm can be fed as input to higher-level processing tasks, such as model-based object recognition systems. As is known, human image perception is based on removing selectively image structures or objects while preserving the other ones. This selection is mainly based on the geometry and local contrast of the images objects [12]. In this sense, any image metric suitable for different filtering tasks can already be interpreted as a step towards the interpretation of the image [23, 18]. Also a key issue is that, perception itself, plays an important role on human image recognition and segmentation tasks, and that this kind of phenomena is not being properly take into account by any of the contemporary methods developed under the Pattern Recognition, Image Processing and Analysis or Mathematical Morphology scientific areas. However, this kind of human feature is being modelled and analysed by *Gestalt* psychology and philosophical systems since, at least 1910 [24]. It is of much interest to follow that this kind of scientific works point out that perception is a product of a synergistic whole effect, i.e. the effect of perception is generated not so much by its individual elements (e.g. human neurones) as by their dynamic interrelation (collective behaviour) - phenomena that can be found easily in many computational paradigms briefly described above, or even in neural network computational models, where data generalisation, $N$ dimensional matrix re-mapping, pattern classification or forecasting abilities are known to be possible. As putted by *Limin Fu* [10], the *intelligence* of a neural network emerges from the collective behaviour of neurones, each of which performs only very limited operations. Even though each individual neurone works slowly, they can still quickly find a solution by working in parallel. This fact can explain why humans can recognise a visual scene faster than a digital computer, while an individual brain cell responds much more slowly than a digital cell in a VLSI (*Very Large Scale Integration*) circuit. Also, this *brain metaphor* suggests how to build an intelligent system which can tolerate faults (fault tolerance) by distributing information redundantly. It would be easier to build a large system in which most of the components work correctly than to build a smaller system in which all components are perfect. Another feature exhibited by the brain is the associative type of memory. The brain naturally associates one thing with another. It can access information based on contents rather than on sequential addresses as in the normal digital computer. The associative, or content-addressable, memory accounts for fast information retrieval and permits partial or approximate matching. The brain seems to be good at managing fuzzy information because of the way its knowledge is represented.

Typically these systems form a structure, configuration, or pattern of physical, biological, sociological, or psychological phenomena, so integrated as to constitute a functional unit with properties not derivable from its parts in summation (i.e. non-linear) - *Gestalt* in one word [13] (the English word more similar is perhaps *system*, *configuration* or *whole*). This synergetic view, derives from the holistic conviction that the whole is more than the sum of its parts and, since the *energy* in a whole cannot exceed the sum of the energies invested in each of its parts (e.g. first law of thermodynamics), that there must therefore be some quantity with respect to which the whole differs from the mere aggregate. This quantity is called synergy and in many artificial life computational systems can be seen as their inherent emergent and *autocatalytic* properties (process well known for instance in many *Reinforcement Learning* models, namely in Q-learning methods often used in autonomous-agents design [15, chapter 13]). The *Gestalt* psychologists proposed an enormous number of laws, which govern the perceptual grouping from *parts* to *whole* [14]. They are mainly related with: proximity, continuity, co-linearity, co-circularity, parallelism, and symmetry. As mentioned by *Kofka* [14], the co-ordination of these rules is guided by the law of *Pragnanz*: "[…] of several geometrically possible organisations that one will actually occur, which possesses the best, simplest and most stable shape". Recent works in Pattern Recognition have applied these principles with success. *Shi* and *Malik* [21] proposed a novel approach for solving the perceptual grouping problem in vision. They treated the image segmentation as a graph-partitioning problem and propose criterions based on the *Gestalt* laws. In other work, *Zhu* [26] proposed a minimax entropy principle for learning probability models through stochastic algorithms for natural images and textures. The shape models are of the form of *Gibbs* distributions defined on *Markov* random fields, whose neighbourhood structures correspond to the *Gestalt* laws. *Zhu* experiments demonstrate that global shape properties can arise from the local interactions of local features. One way to deal with image segmentation, i.e. to find homogeneous partitions in the image is to find their boundaries. In other words, areas with high heterogeneity. In order to model each ant *perception* of heterogeneous areas on the digital image habitat, robust metrics should be introduced. These metrics operate at local neighbourhoods for each ant in the colony, and in some sense they represent the individual "probability matrices" defined by *Wilson* [25], in our understanding of the *mass behaviour*. Then, by pheromone deposition (modelled in [4]), and extending this deposition to be proportional to those correlation values, each individual contributes to the swarm (whole) *perception* of the image, which in turn also serves as their *habitat*. Although the following image correlation metrics are far from being ideal in *Gestaltic* terms, the present and preliminary model made use of them.

### 3.1. Evolutionary Measures for Capturing Heterogeneous Fields

*Bhat* [3] has presented an evolutionary measure for image matching that is based on the *Ulam's* distance - a well know ordinal measure from molecular biology, based on an evolutionary distance metric that is used for comparing real DNA strings. Given two strings, the *Ulam's* distance is the smallest number of mutations, insertions, and deletions that can be made within both strings such that the resulting substrings are identical. Then, *Bhat* reinterprets the *Ulam's* distance with respect to permutations that represent windows intensities expressed on an ordinal scale. The motivation for him to use this measure is twofold: it not only gives a robust measure of correlation between windows but also helps in identifying pixels that contribute to the agreement (or disagreement) between the windows. Given a set of two window $(I_1, I_2)$ intensity values given by $(I^i_1, I^i_2)^n_{i=1}$, let $\pi^i_1$ be the rank of $I^i_1$ among the $I_1$ data, and $\pi^i_2$ be the rank of $I^i_2$ among the $I_2$ data. Consider for instance the following example with two 3 X 3 windows, being $n = 9$. Then $(I^i_1)^n_{i=1}=$:

| 10 | 30 | 70 |
|---|---|---|
| 20 | 50 | 80 |
| 40 | 60 | 100 |

i.e. $(I^i_1)_{i=1}=10$, $(I^i_1)_{i=2}=30$, …, $(I^i_1)_{i=9}=100$. For $(I^i_2)^n_{i=1}$ we have:

| 10 | 30 | 70 |
|---|---|---|
| 20 | 50 | 80 |
| 40 | 60 | 15 |

Thus we have for $\pi^i_1$:

| 1 | 3 | 7 |
|---|---|---|
| 2 | 5 | 8 |
| 4 | 6 | 9 |

and for $\pi^i_2$ the following rank matrix:

| 1 | 4 | 8 |
|---|---|---|
| 3 | 6 | 9 |
| 5 | 7 | 2 |

we then define a composition permutation $s^i$ as:

$$\left(s^i\right)^9_{i=1} = \pi_2^{i:\pi^i_1=1}, \pi_2^{i:\pi^i_1=2}, ..., \pi_2^{i:\pi^i_1=9}$$

i.e., informally $s^i$ is the rank of the pixel in $I_2$ that corresponds to the pixel with rank $i$ in $I_1$. That is, $(s^i)^9_{i=1}=(1,3,4,5,6,7,8,9,2)$. Under perfect positive correlation between two windows, $s^i$ should be identical to the *identity permutation* given by $(u^i)^9_{i=1}=(1,2,3,4,5,6,7,8,9)$. Under perfect negative correlation, i.e. the sequences completely disagree, $s^i$ must be identical to the *reverse identity permutation* given by $(r^i)^9_{i=1}=(9,8,7,6,5,4,3,2,1)$. The *Ulam's* distance $\delta$, between the permutations $s^i$ and $u^i$ is defined as the minimum number of elements that must be removed from each permutation such the resulting subsequences agree perfectly. This is also equal to $n$ minus the length of the longest increasing subsequence in $s^i$. For obtaining the distance between $s^i$ and $r^i$, we construct $s^{*i}$ as:

$$s^{*i} = s^{n-i+1}, i = 1,...,n$$

$$\left(s^{*i}\right)^9_{i=1} = (2,9,8,7,6,5,4,3,1)$$

The *Ulam's* distance between $s^i$ and $r^i$ is equal to that between $s^{*i}$ and $u^i$. For our example we then have the longest common increasing subsequence between $s^i$ and $u^i$ as $(1,3,4,5,6,7,8,9)$ and between $s^{*i}$ and $u^i$ as $(2,9)$. Then we have the *Ulam's* distance between $s^i$ and $u^i$, $\delta_1 = \delta(s^i, u^i) = 9 - 8 = 1$. Similarly, the *Ulam's* distance between $s^{*i}$ and $u^i$, $\delta_2 = \delta(s^{*i}, u^i) = 9 - 2 = 7$. Both $\delta_1$ and $\delta_2$ can take values in the range $[0,n-1]$. Then two measures of correlation $\tau_u$ and $\tau_r$ can be defined as:

$$\tau_u = 1 - \frac{2\delta_1}{n-1} = 0.75; \tau_r = 1 - \frac{2\delta_2}{n-1} = -0.75$$

Both are distributed in the range [-1,1]. To obtain a measure that is symmetrically distributed around zero, we can define an average quantity $\tau$ as:

$$\tau = \frac{\tau_u - \tau_r}{2} = 0.75$$

The previous approach, even if robust, has some drawbacks. The problem is to define the appropriate ranking matrices for tied ranks, which although can be possible within grey level intensities, is more probable to happen in the case of binary images. If this is the case, we must then use another strategy to code the ranking matrices. One possible strategy is to compute the $(\pi^i_1, \pi^i_2)^n_{i=1}$ rankings assuming that the importance for each tied group of cells are redefined by the following order:

| 1 | 2 | 3 |
|---|---|---|
| 4 | 5 | 6 |
| 7 | 8 | 9 |

To deal with the same problem, although not mentioned in [3], *Bhat* and *Nayar* [2] ordered tied pixels in raster scan fashion which is similar to the present strategy. It makes sense as it emphasises positive correlation between corresponding windows. However, it should be noticed that one of the drawbacks of ordinal measures, in general, is that we lose information by going to an ordinal scale. In the case of grey-level images, by viewing intensity as an ordinal variable we lose the grey-level ratio information between pixels.

### 3.2. Statistical Measures

Other developed measure made use of simple statistical computations. In order to measure degrees of similarity between two different lattice windows, in terms of grey level spatial arrangements, a measure $\Delta_h$ was introduced (where *h* stands for *habitat*). $\Delta_h$ possesses three types of terms (weigthed by three constants with sum equal to one), being each term responsible for one type of change. The first term, computed trhough differences in simple averages is responsible for finding differences on grey level overall intensity values, while the second measures differences on windows grey level homogeneity values through variance computations. The last term measures successful matching properties between windows considering all types of permutations. This last term is computed trhough differences in two grey level histograms representative of two local neighbourhoods. Thus;

$$\Delta_h = \left[ a \frac{|m_1 - m_2|}{Max|m_1 - m_2|} + b \frac{|\sigma_1^2 - \sigma_2^2|}{Max|\sigma_1^2 - \sigma_2^2|} + c \frac{S}{S_{max}} \right].(a+b+c)^{-1}$$

where $(a+b+c)=1$. Grey level average intensities in window one are represented by $m_1$, while $\sigma^2_1$ represents the variance for the same window, wich is equal to $(\sum_i g^2_i - n.g^2_{average}).n^{-1}$. On the other hand, $S$ equals to the difference for all grey level intensities ($i=0,...i=255$; for 8 bit images) between two grey level histograms $f(i)$ representative of two windows $w_1$ and $w_2$, i.e. $\sum_i |f_1(i) - f_2(i)|$.

## 4. Extended model

In order to achieve emergent and *autocatalytic* mass behaviours around heterogeneous fields on the *habitat* (i.e. the image), which can significantly change the expected ant colony cognitive map (pheromonal field), instead of a constant pheromone deposition rate $\eta$ (see section 2) used in [4], a term not constant is included. This term is naturally related with the proposed correlation measures around local neighbourhoods, introduced in sections 3.2.1 and 3.2.2. For instance, if we use $\Delta_h$ as that measure, the pheromone deposition rate $T$ for a specific ant at a specific cell (at time $t$), should change to a dynamic value ($p$ is a constant):

$$T = \eta + p\Delta_h$$

Notice that, if the image is completely homogeneous ($Z$ on fig.2 - i.e. if all intensity grey levels are equal) results expected by this extended model will be equal to those found by *Chialvo* and *Millonas* in [4], since $\Delta_h$ equals to zero. In this case, this is equivalent to say that only the swarm pheromonal field is affecting each ant choices, and not the *environment*. I.e. the expected network of trails depends largely on the initial random position of the colony, and in clusters formed in the initial configurations of pheromone, through relative distances. On the other hand, if this environmental term is added, a stable configuration will appear, which is largely independent on the initial conditions of the colony, and becomes more dependent on the *habitat* itself. I.e. the convergence of *habitat perceptive* patterns will occur.

## 5. Results

Twelve tests were conducted, which are related with five 8 bit grey level images with 100 per 100 pixels each. Images are: *A*-Cross (synthetically produced), *B*-Einstein, *C*-Map (scanned from paper), *D*-Marble and *E*-Road. In general (except when indicated), values used for running these experiments were those from [1] ($\eta=0.07$, $k=0.015$, $\beta=3.5$ and $\delta=0.2$), with $p=1.5$, without allowing ants to step each other, and with 3000 individuals ($n=0.3$), i.e. 30% of the squared lattice global area. Finally, one last test was performed. One artificial colony was thrown to the *Einstein habitat* where 1000 iterations were computed. Then at $t=100$, this last habitat was removed and in his place we have introduced the *Map* image. Results are available through figures 2-3, via codification of pheromone fields into 8 bit images. This codification was proportional to the pheromone level at each cell (i.e. = 255, white, if the pheromone level was maximum in the colony, and black if that cell has been completely evaporated in pheromone). One thousand iterations are computed in 2.68 minutes (3000 ants / Pentium II / 333 MHz / 128 MB Ram).

## 6. Conclusions and Future Work

Evolution of mass behaviours on time are difficult to predict, since the global behaviour is the result of many part relations operating in their own local neighborhoud. The emergence of network trails in ant colonies, for instance, are the *product* of several simple and local interactions that can evolve to complex patterns, which in some sense translate a meta-behaviour of that swarm. Moreover, the translation of one kind of low-level (present in a large number) to one meta-level is minimal. Although that behaviour is specified (and somehow constrained), there is minimal specification of the mechanism required to generate that behaviour; global behaviour evolves from the many relations of multiple simple behaviours, without global coordination (i.e. from local interactions to global complexity, like in any *Cellular Automata* model [11]). One paradigmatic and abstract example is the notion, within a specified population, of *common-sense*, being the meta-result a type of *collective-conscience*. There is some evidence that our brain operates in the same way, and as a consequence our perception capabilities derive from emergent properties, which cannot be neglected in any pattern recognition algorithm. These systems show in general, interesting and desirable features as *flexibility* (e.g. the brain is able to cope with incorrect, ambiguous or distorted information, or even to deal with unforseen or new situations without showing abrupt perfomance breakdown) or *versability* quoting *Dorigo* again [6,1], *robustness* (keep functioning even when some parts are locally damaged – see Damásio [7]), and they operate in a massively parallel fashion. Present results point to that type of features. Although this preliminary model is far from being consistent with the *Gestalt* theory of perception, and in many instances sensible to only the most important features of one image (as humans perceive it), swarm pheromonal fields reflect some convergence towards *recognition* of any type of image. There is however one interesting feature. While the swarm is reacting more to the environment than to the initial conditions of pheromone distribution, different local image correlation measures seem to have little importance. Results point to similar consequences using *Ulam's* measure or $\Delta_h$ (proposed in section 3.2). Although relevant to the colony to achieve any perceptive capabilities, their form seems irrelevant. Moreover, the present model shows important adaptive capabilities, as in the presence of sudden changes in the *habitat* (fig. 3). Even if the model is able to quickly adapts to one specific environment (fig. 2), evolving from one empty pheromonal field, *habitat* transitions point that, the whole system is able to have some memory from past environments (i.e. convergence is more difficult after *learning* and *perceiving* one *habitat*). Future work includes three main lines. First, it is our aim to study the effects of every single parameter into the recognisable objects of one image. Secondly, to develop better metrics towards an effective *Gestalt* theory of perceptual grouping. Finally, our aim is to develop several casts of ant colonies, each one responsible of one type of *Mathematical Morphology* operator (MM, *Serra* [22]). As stated in [5], the removal of closure noise (which is described as a fundamental property of pattern grouping as advocated by the *Gestalt* theory, and can provide a key to image understanding in human visual perception) may be done by the *Serra's* formulation of dilation and erosion (the two basic *Mathematical Morphology* operators from which all the others are possible). In this sense, not only ant systems receive feedback from any digital image *habitat*, as they can even change the *habitat* accordingly. It is our belief that the present work may lead to a new way to develop more powerful computational perceptive systems, towards an distributed, adaptive and bio-inspired *Mathematical Morphology*, aiming to design MM operators as artificial living organisms. We wish that the present work could stimulate better research results along this direction.

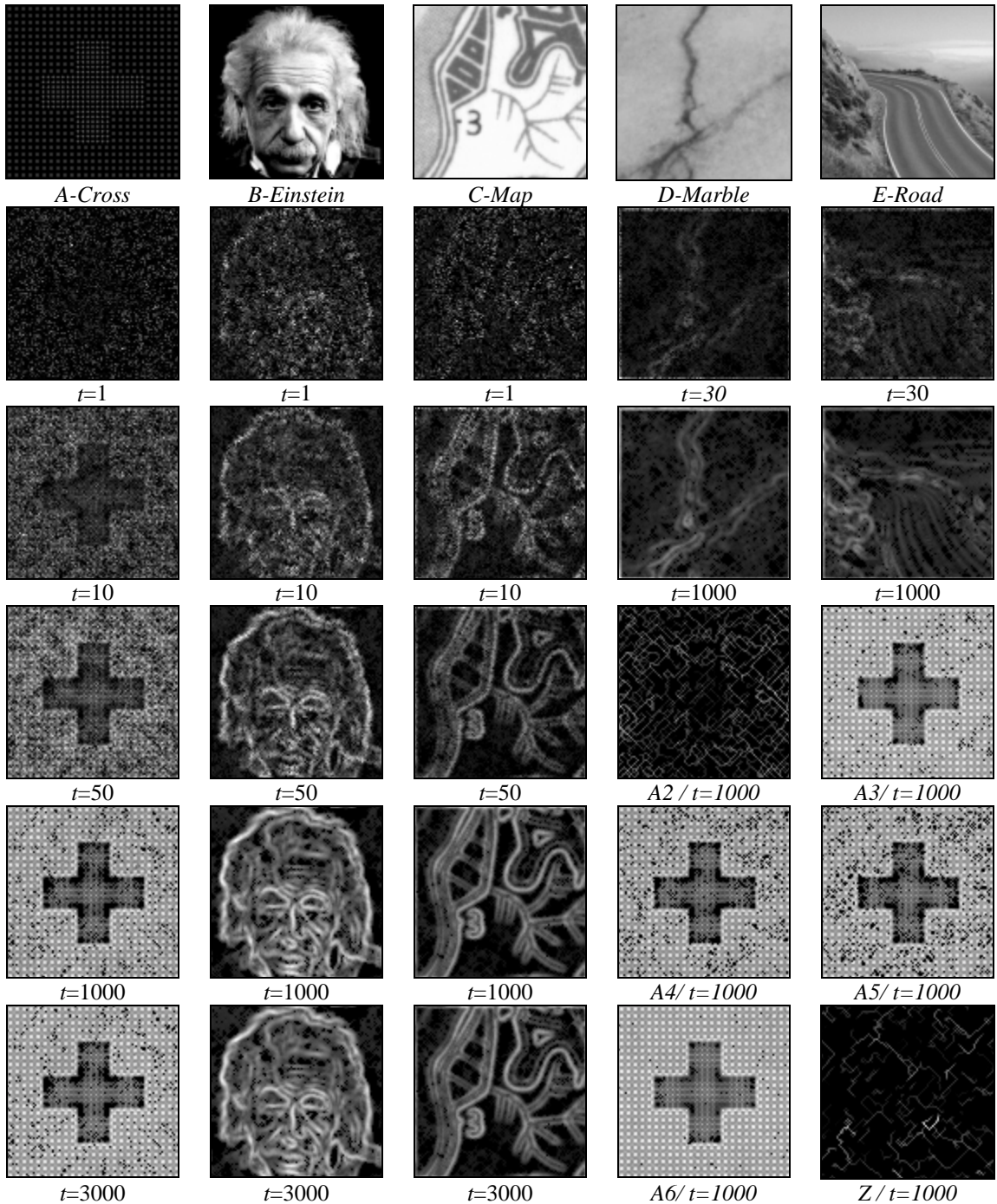

Figure 2 - Colony cognitive maps (pheromonal fields) for several iterations, on images *Cross*, *Einstein*, *Map*, *Marble* and *Road*. Except when indicated, parameters are those from [4] (see section 5). In *A2* and *Z*, ants are allowed to step on each other; *habitats* are respectively *Cross* and an homogeneous image. In this last case, results are similar with those found by *Chialvo* and *Millonas* [4]. *A3*) $k=0.011$. *A4*) $k=0.019$. *A5*) $\beta=4.5$. *A6*) $\beta=2.5$.

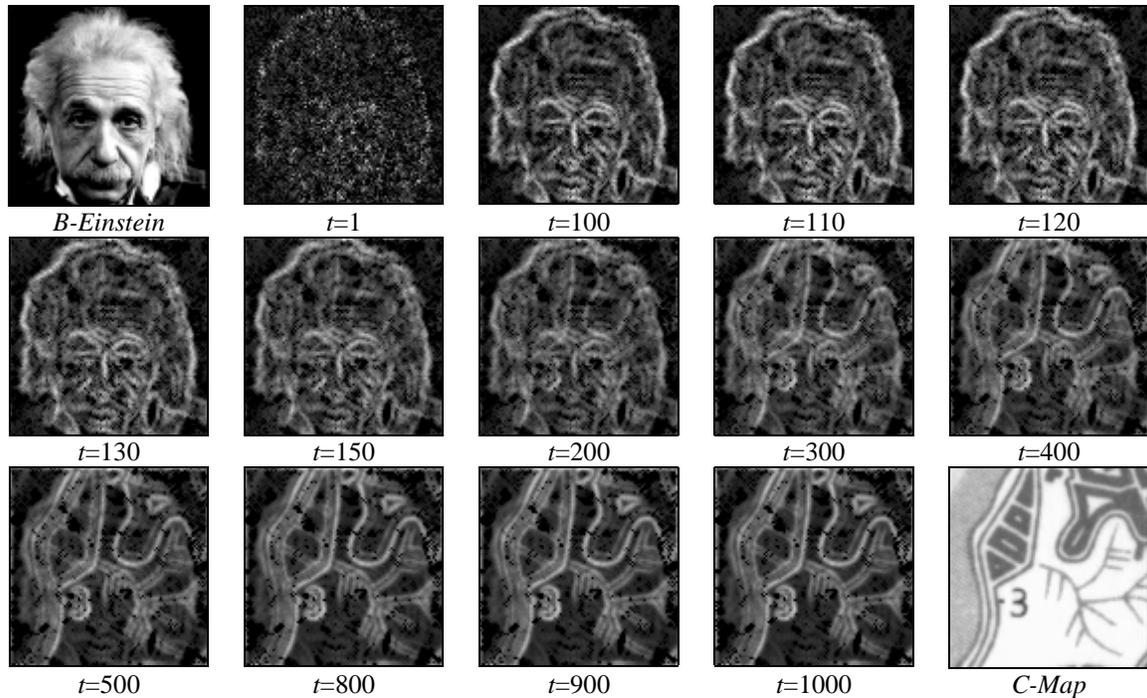

Figure 3 - One swarm (3000 ants) is thrown to explore *Einstein* image for 1000 iterations. At t=100, the *Einstein* habitat is replaced by *Map* image. Evolution of swarm cognitive maps (pheromonal fields) are shown for several iterations.

**Acknowledgements:** Both authors wishes to thank to *Manuel Gomes* (VARIOGRAMA) for his helpful advises on *Delphi* Computer Graphics environments and code. The first author, wishes also to thank to FCT-PRAXIS XXI (BD20001-99), for his PhD Research Fellowship.